\documentclass[lettersize,journal]{IEEEtran}
\usepackage{amsmath,amsfonts,amssymb}
\usepackage{algorithm}
\usepackage{array}
\usepackage[caption=false,font=normalsize,labelfont=sf,textfont=sf]{subfig}
\usepackage{textcomp}
\usepackage{stfloats}
\usepackage{url}
\usepackage{verbatim}
\usepackage{graphicx}
\usepackage{multirow}
\usepackage{cite}
\usepackage{algpseudocode}
\usepackage{bm}

\usepackage[table]{xcolor}
\usepackage{booktabs,makecell}
\usepackage{xcolor}
\usepackage{bbding}
\usepackage{pifont}
\usepackage{hyperref}

\begin{document}

\title{Collaborative-Distilled Diffusion Models (CDDM) for Accelerated and Lightweight Trajectory Prediction}

\author{Bingzhang Wang, 
        Kehua Chen,~\IEEEmembership{Member,~IEEE}, 
        and Yinhai Wang,~\IEEEmembership{Fellow,~IEEE}

\thanks{Bingzhang Wang, Kehua Chen, and Yinhai Wang are with the Department of Civil and Environmental Engineering, University of Washington, Seattle, WA 98195. (E-mail: \texttt{\{bzwang,zeonchen,yinhai\}@uw.edu})}
}



\maketitle

\begin{abstract}
Trajectory prediction is a fundamental task in Autonomous Vehicles (AVs) and Intelligent Transportation Systems (ITS), supporting efficient motion planning and real-time traffic safety management. Diffusion models have recently demonstrated strong performance in probabilistic trajectory prediction, but their large model size and slow sampling process hinder real-world deployment. This paper proposes Collaborative-Distilled Diffusion Models (CDDM), a novel method for real-time and lightweight trajectory prediction. Built upon Collaborative Progressive Distillation (CPD), CDDM progressively transfers knowledge from a high-capacity teacher diffusion model to a lightweight student model, jointly reducing both the number of sampling steps and the model size across distillation iterations. A dual-signal regularized distillation loss is further introduced to incorporate guidance from both the teacher and ground-truth data, mitigating potential overfitting and ensuring robust performance. Extensive experiments on the ETH-UCY pedestrian benchmark and the nuScenes vehicle benchmark demonstrate that CDDM achieves state-of-the-art prediction accuracy. The well-distilled CDDM retains \emph{96.2\%} and \emph{95.5\%} of the baseline model's ADE and FDE performance on pedestrian trajectories, while requiring only \emph{231K} parameters and \emph{4} or \emph{2} sampling steps, corresponding to \emph{161$\times$} compression, \emph{31$\times$} acceleration, and \emph{9 ms} latency. Qualitative results further show that CDDM generates diverse and accurate trajectories under dynamic agent behaviors and complex social interactions. By bridging high-performing generative models with practical deployment constraints, CDDM enables resource-efficient probabilistic prediction for AVs and ITS. Code is available at \href{https://github.com/bingzhangw/CDDM}{https://github.com/bingzhangw/CDDM}.
\end{abstract}


\section{Introduction}
As the rapid development of Autonomous Vehicles (AVs) and Intelligent Transportation Systems (ITS), an increasing trend of research advancement in trajectory prediction has emerged. Trajectory prediction refers to the predictive estimation of traffic agents' future motion or states (e.g., vehicles, pedestrians) in complex surrounding environments. It is both a challenging task and a fundamental technology for enabling reliable autonomous system pipelines. For AVs, effective maneuvering in heterogeneous real-world environments with diverse traffic participants relies on predictive trajectory estimation, which supports safe and efficient motion planning~\cite{tpreviewav, humanlike, highorder}. For ITS, fast and resilient trajectory prediction is critical to real-time safety operations on roadside infrastructures, especially in scenarios where collisions may occur within seconds between high-speed vehicles and vulnerable road users under dynamic traffic conditions at urban intersections or highways~\cite{tpvru, accidentprediction}.

\begin{figure}[t]
    \centering
    \includegraphics[width=0.95\columnwidth]{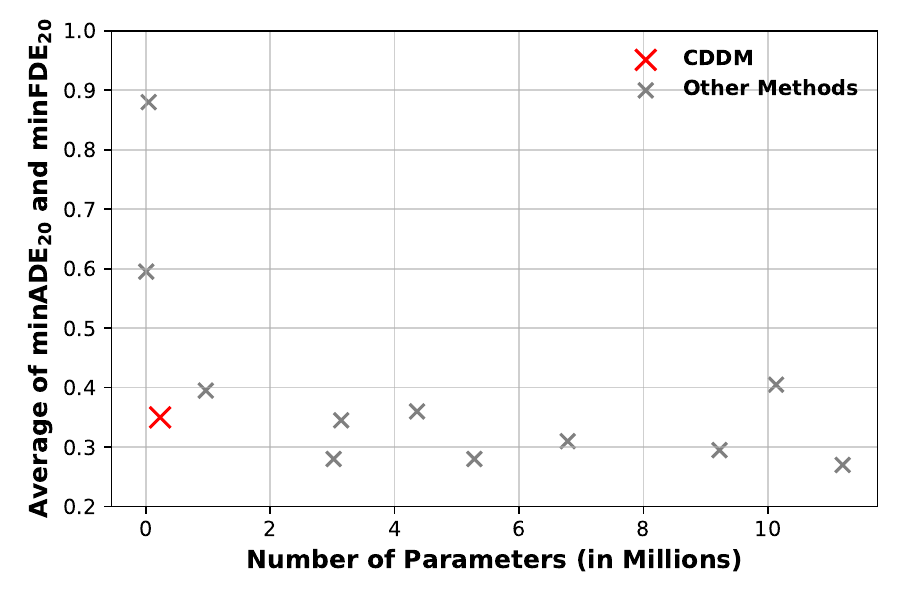}
    \caption{Comparison of average $\min ADE_{20}$ and $\min FDE_{20}$ with respect to model size on the ETH-UCY benchmark. CDDM (red) achieves state-of-the-art trajectory prediction error while using substantially lower computational cost than other baseline methods in recent years (gray).}
    \label{fig:modelsize}
\end{figure}

In response to this pressing need from the industrial and academic sectors, a significant amount of research has been conducted to develop accurate and reliable trajectory prediction algorithms and models~\cite{tpreviewav, pedsbehaivor, humanmotiontp, tpvru}. To precisely capture the historical information, prior studies have developed methods to model complex interactions among heterogeneous traffic agents~\cite{robusttp}, and to encode influential factors such as surrounding environments, ego-vehicle states and high-definition map rasterization~\cite{phan2020covernet}. Generally, these components are referred to encoders. A group of deep learning models has shown their superior capabilities as encoders, such as Graph Neural Networks~\cite{pmlr-v164-deo22a, YuMa2020Spatio}, Convolutional Neural Networks~\cite{multimodaltpconvnet, vrutpconvnet, traphic}, and Attention Mechanism~\cite{GAT, GRANP, aitp}. Subsequently, to convert the learned contextual feature embeddings into future motions like velocities, headings, or positions, various research have proposed decoders based on time-series forecasting models, such as Transformer~\cite{thermalforecast, GAT, YuMa2020Spatio} and Recurrent Neural Network family, including Gated Recurrent Units~\cite{fusiongru}, Long Short Term Memory~\cite{sociallstm, lstmhighway}, and so on. These deterministic trajectory prediction methods, although having achieved a comparable performance, are able to generate only unimodal trajectory predictions. In simple terms, it means making a single best guess of the future trajectory. Such an attribute of deterministic prediction methods conflicts with the nature of traffic agent motion, which implicitly involves nondeterministic intentions or goals~\cite{densetnt}. Hence, these methods constrain the model's ability to capture uncertainty in motion prediction~\cite{uncertainty}. 

In the most recent years, probabilistic prediction methods for multi-modal trajectory distribution learning using deep generative models~\cite{genai}, such as Generative Adversarial Networks~\cite{goalgan}, Conditional Variational Auto-Encoders~\cite{trajectron++}, Diffusion Models~\cite{mid}, and Large Language Models~\cite{lcllm}, have emerged as innovative solutions to countermeasure the above-mentioned bottleneck. Among generative models, Diffusion Models and their variants have been experimented to be able to achieve the top performance in trajectory prediction tasks~\cite{diffusionview, diffusiontrans}, mainly due to their multiple steps of denoising process to generate high-fidelity predictions. However, trajectory prediction in AV and ITS domains typically needs to be performed efficiently on edge devices~\cite{stmgcn, prednet}. 
The slow sampling process and large model size hinder the real-world application of the Diffusion Models in trajectory prediction context~\cite{diffusiontrans}.




For countermeasures, several diffusion acceleration methods are introduced to improve inference speed \cite{ddim, consistency, dydit, progressive}. Nonetheless, these methods either strive to address diffusion sampling acceleration or leverage conventional model compression techniques to shrink backbone model size through knowledge distillation, neural network pruning, or quantization. To the best of our knowledge, none of the prior research has effectively achieved both goals of Diffusion Model acceleration and compression. To tackle this significant gap in the context of accurate trajectory prediction, this paper proposes Collaborative-Distilled Diffusion Models (CDDM) for efficient, cost-effective, lightweight, real-time trajectory prediction that only requires as few as two to four sampling steps for accurate prediction of pedestrian and vehicle trajectories. Experiments on open-sourced real-world datasets show that our proposed method significantly reduced the computational cost (FLOPs) during diffusion sampling and effectively integrated a compressed, lightweight (number of parameters) transformer-based model as diffusion backbone, while still achieving the state-of-the-art performance among other recent baseline methods (as shown in Figure~\ref{fig:modelsize}). This paper highlights the following major contributions:
\begin{itemize}
    \item We propose Collaborative-Distilled Diffusion Models (CDDM) for real-time and lightweight pedestrian and vehicle trajectory prediction. CDDM is built upon Collaborative Progressive Distillation (CPD), a novel two-stage knowledge distillation framework that achieves both diffusion model compression and sampling step acceleration in a collaborative process. Unlike prior approaches that focus on either compression or acceleration in isolation, CDDM progressively distills a lightweight student diffusion model from a larger teacher model while simultaneously halving the number of denoising steps, enabling efficient, scalable, and accurate trajectory prediction.
    \item Our framework introduces a regularized diffusion distillation loss that integrates signals from both the teacher model and the ground-truth data. This dual-signal design mitigates overfitting to suboptimal or biased teacher predictions and ensures consistent performance as model capacity and sampling steps are progressively reduced across distillation iterations, while effectively transferring the teacher’s knowledge to the student.
    \item We conduct extensive experiments on the ETH-UCY pedestrian benchmark and the nuScenes vehicle benchmark, showing that our CDDM, distilled using a lightweight model with \emph{231K} parameters, achieves state-of-the-art performance with only \emph{4} or \emph{2} sampling steps. The \emph{4}-step lightweight CDDM retains \emph{96.2\%} and \emph{95.5\%} of the baseline model’s ADE and FDE performance on pedestrian trajectories, while requiring only \emph{0.62\%} of the model size (\emph{161$\times$} compression) and \emph{3.2\%} of the inference time (\emph{9 ms} latency, \emph{31$\times$} acceleration).
\end{itemize}

The remainder of this paper is organized as follows. Section~\ref{sec:relatedwork} reviews related work. Section~\ref{sec:preliminaries} formulates the preliminaries of our approach. The proposed CDDM methodology is described in detail in Section~\ref{sec:CDDM}. Section~\ref{sec:experiment} reports experimental results, and Section~\ref{sec:conclusion} concludes the paper.

\section{Related Work}
\label{sec:relatedwork}
Trajectory prediction methods can be generally classified into categories based on different technical approaches used, such as physics-based methods, classic machine learning-based methods, deep-learning based methods, and reinforcement learning-based methods~\cite{tpreviewav}. In this section, we summarize the most recent research from a different perspective of whether indeterminacy forecasting is integrated in multi-modal trajectory prediction, thus emphasizing on two categories: classic deterministic deep learning methods, and probabilistic deep generative methods. Thereafter, we review innovative efficient Diffusion Models for sampling acceleration methods.

\subsection{Deterministic Trajectory Prediction}
Representing most of classic prediction approaches, deterministic trajectory prediction methods commonly output uni-modal results without forecasting probability scores, considered as the single best-guess prediction of the future trajectory. Models under this category have demonstrated impressive performance primarily due to their stability and efficiency. 

As a pioneering deep-learning approach for predicting human trajectories, Alahi et al. \cite{sociallstm} presented Social LSTM to model social interactions among pedestrians in crowded environments. A sequence-to-sequence architecture is proposed to effectively capture individual dynamics and the influence of nearby agents. Subsequently, trajectory prediction in dense and heterogeneous traffic environments became the research focus that differs from conventional homogeneous methods. Chandra et al. \cite{traphic} proposed TraPHic to leverage LSTM-CNN network to account for horizon-based interactions among heterogeneous road agents of varying shapes, dynamics and behaviors. Graph-based methods were also widely applied to modeling spatio-temporal interactions among traffic agents, especially by leveraging attentive modules to extract intentions. Yu et al. \cite{YuMa2020Spatio} proposed STAR, a spatio-temporal graph transformer framework to model intra-graph crowd interaction by only using attention mechanisms. Their separate spatial and temporal transformers are capable of capturing complex socially aware spatio-temporal dependencies of crowd. Similarly, Zhang et al. \cite{GAT} proposed a Graph Attention Transformer to further model agent-agent and agent-infrastructure interactions in sparse-graph represented traffic scenes. In the mean time, rasterizing local map into colorized raster image has become a prevalent method to enhance contextual understanding in trajectory prediction. The visual features from encoded raster images can indicate surrounding agent locations, lane layout, trajectory path, infrastructure setup, and other environmental information, which acts as the input for prediction backbones typically CNN-based models. Chou et al. \cite{vrutpconvnet} presented an efficient architecture to rasterize high-definition maps and actor's surroundings into bird-eye view images used as the input to convolutional networks, which demonstrated a favorable performance balance between prediction accuracy and computation efficiency. Alternatively, end-to-end trajectory prediction directly based on visual inputs and detection bounding boxes emerged as a practical solution that takes into account original video feeds or sensing data. As representative work, Karim et al. \cite{thermalforecast} proposed ViT-DGRU architecture for robust trajectory prediction under challenging nighttime low-visibility conditions, which inputs thermal video as visual aids and bounding box trajectories as temporal dynamics for global scene feature extraction. 

Nevertheless, although achieving comparable performance, these deterministic trajectory methods struggle to capture motion uncertainty or stochastic intention from the nature of multi-modal traffic-agent movements.

\subsection{Stochastic Generative Models}
The emergence of various stochastic generative models has provided opportunities for the employment of probabilistic prediction within a broad range of applications~\cite{genai}, such as computer vision, natural language processing~\cite{tpgpt}, temporal data modeling, and so on. These probabilistic models explicitly predict the data distribution normally using multivariate Gaussian distribution, from which multi-modal outputs are then sampled during the inference time. A fair number of research has exhibited strong applicability of these deep generative models to challenging real-world downstream tasks, particularly trajectory prediction in this study, due to their flexibility and strength. Among diverse generative models, Diffusion Models \cite{ddpm}, Variational Autoencoders (VAEs) \cite{vae}, and Generative Adversarial Networks (GANs) \cite{gan} are the most representative approaches adopted for probabilistic trajectory prediction and are therefore the focus of this section. 

Ho et al. \cite{ddpm} proposed Denoising Diffusion Probabilistic Models (DDPM), the original and widely used diffusion model framework. DDPM consists of a multi-step forward diffusion process that incrementally adds stochastic noise to original inputs until close to standard Gaussian noise, and a reverse denoising process that learns to predict and remove added noise of each step from the corrupted data. Gu et al. \cite{mid} firstly applied DDPM to stochastic trajectory prediction by proposing Motion Indeterminacy Diffusion (MID), which explicitly simulates the progress of discarding indeterminacy from all the potential movement areas until reaching the desired trajectory. MID uses a Transformer-based architecture as the diffusion backbone to predict Gaussian noises in reverse denoising process conditioned on contextual encodings, and has achieved optimal performance on public pedestrian trajectory benchmarks compared to other state-of-the-art methods. Afterwards, Chen et al.  \cite{equidiff} incorporates an enhanced backbone model in their Conditional Equivariant Diffusion Model (EquiDiff), by proposing an SO(2)-equivariant transformer that fully utilizes the geometric properties of location coordinates. 

In the domain of time series forecasting, Conditional Variational Autoencoders (CVAEs) \cite{cvae} as a variant of VAEs, are more commonly adopted due to their ability to condition the prediction on contextual information, likely including historical inputs, social interactions, or environmental factors. Trajectron++ proposed by Salzmann et al. \cite{trajectron++} is often present as the baseline method among a set of CVAE-based trajectory prediction methods. In their model architecture, driving scenes are represented as a directed spatio-temporal graph to model agent interactions learned by encoders, and then multi-modal trajectories are generated by leveraging the discrete latent variable framework of CVAE to model high-level categorical behaviors. Additionally, dynamics are integrated with the predicted control actions (e.g., acceleration and steering rate) to produce dynamically-feasible trajectories. 

GAN-based methods utilize an adversarial process to implicitly capture data distributions, but require careful training to prevent mode collapse. Multi-Agent Tensor Fusion (MATF) presented by Zhao et al. \cite{MATF} is one of the best-performing model under this category. MATF employs a conditional GAN, where the generator forecasts future trajectories for all agents based on past movements, scene context, and random noise, and the discriminator leans to distinguish real from generated trajectories. The adversarial training facilitates generation of diverse and realistic trajectories. 

Overall, research employing stochastic generative models has shown a significant advancement towards more diverse, accurate, and uncertainty-aware trajectory prediction. Among the models discussed, Diffusion Models and variants particularly demonstrate superior performance in both training and inference, owing to their training stability and high-quality sampling results. However, the relatively slow inference speed, due to required multi-step denoising process, limits their applicability in scenarios that demand rapid response. 


\subsection{Efficient Diffusion Models}
\label{sec:efficient}

Although Diffusion Models have demonstrated remarkable capabilities in high-quality generation tasks, they require simulating a Markov chain in the denoising process to produce a sample, which has hindered their further applications due to long sampling steps an high computational costs. In order to improve sampling efficiency, a number of studies have explored Diffusion Models acceleration methods that can be generally categorized into three paradigms: algorithmic acceleration, knowledge distillation, and architecture improvement.

Algorithmic acceleration methods usually reduce sampling steps by reformulating the diffusion process without the need to retrain backbone models. Song et al. \cite{ddim} presented Denoising Diffusion Implicit Models (DDIM) to generalize DDPM to a non-Markovian process that shares the same training objective. This non-Markovian process of DDIM can correspond to a deterministic generative process that enables to skip arbitrary number of steps in sampling process. DDIM shows the ability to produce high-quality experimental samples 10$\times$ to 50$\times$ faster than DDPM.

Knowledge distillation-based methods typically train a student model to replicate the inference results of a multi-step teacher model, but with significantly fewer steps and lower computational cost \cite{diffkdsurvey}. Salimans and Ho \cite{progressive} proposed progressive distillation for fast sampling. Their method distills a trained DDIM diffusion sampler using many steps, into a new diffusion model that takes half as many sampling steps. They then keep progressively applying this distillation procedure to halve the number of sampling steps each time, until the model achieves comparable performance in only four steps. Song et al. \cite{consistency} proposed Consistency Models that support one-step generation by design, while still allowing multi-step sampling. The model can be trained either by distilling pretrained diffusion models, or as standalone generative models.

Architecture-based methods optimize diffusion model architecture to reduce inference computation or achieve fast few-step sampling. Zhao et al. \cite{dydit} proposed Dynamic Diffusion Transformer (DyDiT), a novel architecture that dynamically adjusts computation along timestep and spatial dimensions during generation, which significantly reduces computational operations in inference time. Mao et al. \cite{leapfrog} proposed an accelerated diffusion-based trajectory prediction method, Leapfrog Diffusion Model (LED), which incorporates a trainable leapforg initializer to learn a multi-model distribution of future trajectories, enabling the model to skip a large number of denoising steps. Other methods use lightweight neural networks to accelerate inference process, or exploit model pruning, quantization, or knowledge distillation to reduce computation.

These discussed approaches aim to either address diffusion sampling acceleration, or leverage conventional model compression techniques to reduce backbone model size; While, none of the research has effectively achieved both goals of Diffusion Models acceleration and compression at the same time.


\section{Preliminaries}
\label{sec:preliminaries}
\subsection{Trajectory Prediction Formulation}
The trajectory prediction task can be formulated as estimating the future positions of traffic participants based on observations of their historical states and surrounding contextual information, which can be denoted as:
\begin{equation}
    \bm{Y}=\mathcal{F}(\bm{X},\bm{C})
\end{equation}
where the input $\bm{X}$ represents historical states that can include position, velocity, acceleration, and heading; $\bm{C}$ represents contextual information that can encode agent interactions, high-definition maps, and environmental factors. For each agent $i$, the input is denoted as:
\begin{align}
    \bm{x}^i = \left\{ \bm{s}_t^i \in \mathbb{R}^2 \;\middle|\; t = -T_\text{hist}+1, -T_\text{hist}+2, \ldots, 0 \right\}, \notag \\
    \forall i \in \left\{ 1, 2, \ldots, N \right\}
\end{align}
where $\bm{x}_i\in\bm{X}$, $\bm{s}_t^i$ denotes the 2D historical states from observations of agent $i$ at timestep $t$, $T_{\text{hist}}$ is the length of the observed trajectory, and current timestep is $t=0$. 

The output $\bm{Y}$ represents the predicted future trajectories. For each agent of interest, it is denoted as:
\begin{equation}
    \bm{y}^i=\left\{\bm{p}_t^i \in \mathbb{R}^2\mid t = 1, 2, \cdots, T_\text{pred} \right\}
\end{equation}
where $\bm{y}_i \in \bm{Y}$, $\bm{p}_t^i$ denotes the predicted 2D position of agent $i$ at timestep $t$, and $T_{\text{pred}}$ is the prediction horizon. $\bm{Y}$ can be obtained either by directly predicting the coordinates at each timestep or by integrating predicted dynamic attributes into kinematic models to generate more physically feasible trajectories.

In contrast to deterministic methods, stochastic trajectory prediction models approximate the distribution of possible future trajectories rather than producing a single output. Individual trajectories are then sampled from this learned distribution. This process can be denoted as sampling from a conditional distribution:


\begin{equation}
\hat{\bm{y}}^i\sim P_\theta\left(\bm{p}_{1:T_\text{pred}}^i \mid \bm{x}^i, \bm{C}  \right)
\end{equation}
where $\hat{\bm{y}}^i$ is the predicted trajectory of agent $i$ and $\theta$ is the parameter of the prediction model. In the following sections, we omit superscript $i$ for simplicity.

Cost-efficient trajectory prediction in the context of limited computational resources, such as deployment on edge-computing devices, requires lightweight model parameters. Therefore, the training objective can be formulated as a constrained optimization problem:

\begin{equation}
\begin{aligned}
    \hat{\theta} =\ & \arg\max_{\theta}\; P_\theta\left(\bm{Y} \mid \bm{X}, \bm{C} \right) \\
    &\text{subject to} \quad \|\theta\| \leq \theta_{\text{constraint}}
\end{aligned}
\end{equation}
where $\hat{\theta}$ denotes the optimal model parameters, $\bm{X}$ and $\bm{Y}$ represent the ground truth input and output, and $\theta_{\text{constraint}}$ specifies the parameter constraint reflecting resource limitations.

\subsection{Denoising Diffusion Probabilistic Models}
\label{sec:ddpm}
The Denoising Diffusion Probabilistic Models (DDPM) \cite{ddpm} have become one of the most widely adopted generative models due to its powerful expressivity. The model consists of a forward process, which gradually adds random noise to the original data without any learnable parameters, and a reverse process, which employs a neural network to predict and remove the added noise from the noised data in multiple steps. Both forward and reverse processes are formulated as a Markov chain with Gaussian transitions. DDPM has also demonstrated strong performance in highly diverse probabilistic trajectory prediction tasks, as introduced by the Motion Indeterminacy Diffusion (MID) \cite{mid}. The following part will introduce how DDPM is applied in the context of trajectory prediction.

First, we formulate the forward trajectory diffusion process $q$:
\begin{equation}
    q(\bm{y}_k \mid \bm{y}_{k-1}) = \mathcal{N}\left(\bm{y}_k; \sqrt{1 - \beta_k} \, \bm{y}_{k-1}, \beta_k \bm{I}\right)  
\end{equation}
where $\bm{y}_k$ denotes the noised future trajectory at diffusion timestep $k$, and $\beta_k$ is the noise schedule parameter, which is typically chosen to follow a linear scheduler in trajectory prediction tasks. Note that our study generalizes the discrete sampling timesteps to a continuous variable defined over the interval $[0, 1]$, to ensure the noise schedule evenly aligned between models with different samping steps. Using the property of Gaussian transitions, we can derive a closed-form expression for the distribution of $\bm{y}_k$ conditioned directly on the original data $\bm{y}_0$:
\begin{equation}
    q(\bm{y}_k \mid \bm{y}_0) = \mathcal{N}\left(\bm{y}_k; \sqrt{\bar{\alpha}_k}\, \bm{y}_0, (1 - \bar{\alpha}_k)\bm{I}\right)
\end{equation}
where $\alpha_k = 1 - \beta_k \quad$ and $\quad \bar{\alpha}_k = \prod_{s=1}^k \alpha_s$. This important property simplifies the training process of diffusion models by allowing to train at a single random timestep for each iteration, without the need to sequentially follow the Markov chain. Meanwhile, when the total diffusion timestep $K$ is large enough, the trajectory data can be approximately considered to follow a standard Gaussian noise distribution as $\bm{y}_K\sim \mathcal{N}(\bm{0}, \bm{I})$.

Next, we formulate the reverse trajectory generation process, which starts from the noise distribution. Unlike the vanilla DDPM, we model the reverse process using a conditional diffusion model where the generation is conditioned on historical agent observations $\bm{X}$ and contextual information $\bm{C}$. A spatio-temporal encoder $\mathcal{F_\varphi}$ parameterized by $\varphi$ is utilized to learn the encoded feature representation $\bm{f}$:
\begin{equation}
    \bm{f} = \mathcal{F}_\varphi(\bm{X}, \bm{C})
\end{equation}
Then, the reverse process can be denoted as:
\begin{equation}
    p_\theta(\bm{y}_{k-1} \mid \bm{y}_k, \bm{f}) = \mathcal{N}\left(\bm{y}_{k-1}; \mu_\theta(\bm{y}_k, \beta_k, \bm{f}),\, \beta_k \bm{I}\right)
\end{equation}
where $\beta_k$ is the predefined noise schedule parameter, and $\theta$ denotes the learnable parameters of the diffusion backbone model. Note that the reverse process begins by sampling $\bm{y}_K$ from a standard Gaussian distribution, and then uses the prediction model to infer the trajectory distribution at the next step in the multi-step Markov chain, iterating until the original timestep is reached. The encoder parameters $\varphi$ and decoder parameters $\theta$ are jointly trained.

\subsection{Denoising Diffusion Implicit Models}
Denoising Diffusion Implicit Models (DDIM) \cite{ddim} accelerate sampling by generalizing DDPM to a non-Markovian diffusion process, which can correspond to a deterministic generation process. DDIM does not require additional training compared to DDPM, since both models share the same training objective. Specifically, DDIM defines the posterior distribution directly by giving:
\begin{align*}
q_{\sigma}(\bm{y}_{k-1} \mid \bm{y}_k, \bm{y}_0) & =\ \mathcal{N} \Bigg( 
\sqrt{\alpha_{k-1}} \bm{y}_0 \\
& + \sqrt{1 - \alpha_{k-1} - \sigma_k^2}
    \frac{\bm{y}_k - \sqrt{\alpha_k} \bm{y}_0}{\sqrt{1 - \alpha_k}},
    \ \sigma_k^2 \bm{I} \Bigg)
\end{align*}
where $\sigma \in \mathbb{R}^K_{\geq0}$ is a predefined variance parameter, which leads to a deterministic inference when $\sigma\rightarrow0$. All other parameters remain the same as DDPM. The mean function is designed to ensure that $q_\sigma(\bm{y}_k|\bm{y}_0) = \mathcal{N}\left( \sqrt{\alpha_k} \bm{y}_0, (1 - \alpha_k) \bm{I} \right)$, so as to maintain the same training process as DDPM. By using Bayes' rule, the non-Markovian forward process can be derived from this formulation, given that $\bm{y}_k$ could depend on both $\bm{y}_{k-1}$ and $\bm{y}_0$.

Now that the forward process no longer follows a Markovian chain, we can define the reverse generation process as a sampling trajectory on a subset of timesteps $\{ \bm{y}_{\tau_1}, \ldots, \bm{y}_{\tau_S} \}$, where $\tau$ is a sub-sequence of $[1,\cdots,K]$ of length $S$ and $S \ll K$:

\begin{equation}
\begin{aligned}
q_{\sigma, \tau}(\bm{y}_{\tau_{i-1}} \mid \bm{y}_{\tau_i}, \bm{y}_0) = \ &
\mathcal{N} \Bigg( \sqrt{\alpha_{\tau_{i-1}}}\bm{y}_0 \\ + 
\sqrt{1 - \alpha_{\tau_{i-1}} - \sigma_{\tau_i}^2} & \cdot
\frac{\bm{y}_{\tau_i} - \sqrt{\alpha_{\tau_i}} \bm{y}_0}{\sqrt{1 - \alpha_{\tau_i}}}, \sigma_{\tau_i}^2 \bm{I}
\Bigg) \quad \forall i \in [S]
\end{aligned}
\end{equation}

Then, we can leverage this re-parameterized posterior to produce samples following customized timestep schedule by calculating:
\begin{multline}
p_\theta^{(\tau_i)}(\bm{y}_{\tau_{i-1}} \mid \bm{y}_{\tau_i}) = \\
\begin{cases}
q_{\sigma, \tau}(\bm{y}_{\tau_{i-1}} \mid \bm{y}_{\tau_i}, \mathcal{F}_\theta^{(\tau_i)}(\bm{y}_{\tau_{i-1}})) 
& \text{if } i \in [S],\; i > 1 \\
\mathcal{N}(\mathcal{F}_\theta^{(t)}(\bm{y}_k),\; \sigma_k^2 \bm{I}) 
& \text{otherwise}
\end{cases}
\end{multline}
where $\theta$ is the learnable parameter of the diffusion backbone neural network. Thus, when the length of the sampling trajectory $S$ is much smaller than $K$, DDIM can significantly improve computational efficiency by reducing the number of sampling iterations, thereby accelerating inference without requiring additional training.

\subsection{Diffusion Model $v$-prediction Parametrization}
\label{sec:v-pre}
Our diffusion model uses the linear noise schedule $\alpha_k=\alpha_1+\frac{k-1}{K-1}(\alpha_K-\alpha_1)$. In trajectory prediction, while most of denoising models aiming at predicting noise $\epsilon$ perform well, they are not suited for distillation because the shortened sampling steps lead to low signal-to-noise ratios (SNR), expressed as $\alpha_k^2/\sigma_k^2$. Therefore, we parameterize our diffusion model to predict the velocity, following the formulation introduced in~\cite{progressive}, defined as:
\begin{equation}
    \bm{v} = \alpha_k \boldsymbol{\epsilon} - \sigma_k \bm{y}_0
\end{equation}
where $\bm{y}_0$ is original data, $\boldsymbol{\epsilon}$ is random noise, $\alpha_k$ and $\sigma_k$ are noise schedule coefficients. The $v$-prediction establishes a connection between data prediction ($x_0$-prediction) and noise prediction ($\boldsymbol{\epsilon}$-prediction), maintaining informative gradients even under low SNR. After further derivation, we can get:
\begin{equation}
\hat{\bm{y}}_0 = \alpha_k \bm{y}_k - \sigma_k \hat{\bm{v}}_{\theta}
\end{equation}
where $\hat{\bm{y}}_0$ denotes the reconstructed data, $\bm{y}_k$ is the noised input at timestep $k$, and $\hat{\bm{v}}_{\theta}$ is the predicted velocity by the backbone neural network, defined as $\hat{\bm{v}}_{\theta} = \mathcal{F}_\theta(\bm{y}_{k}, \beta_{k}, \bm{f})$.

\section{Collaborative-Distilled Diffusion Models for Efficient Trajectory Prediction}
\label{sec:CDDM}

\begin{figure*}[t]
    \centering
    \includegraphics[width=0.95\linewidth]{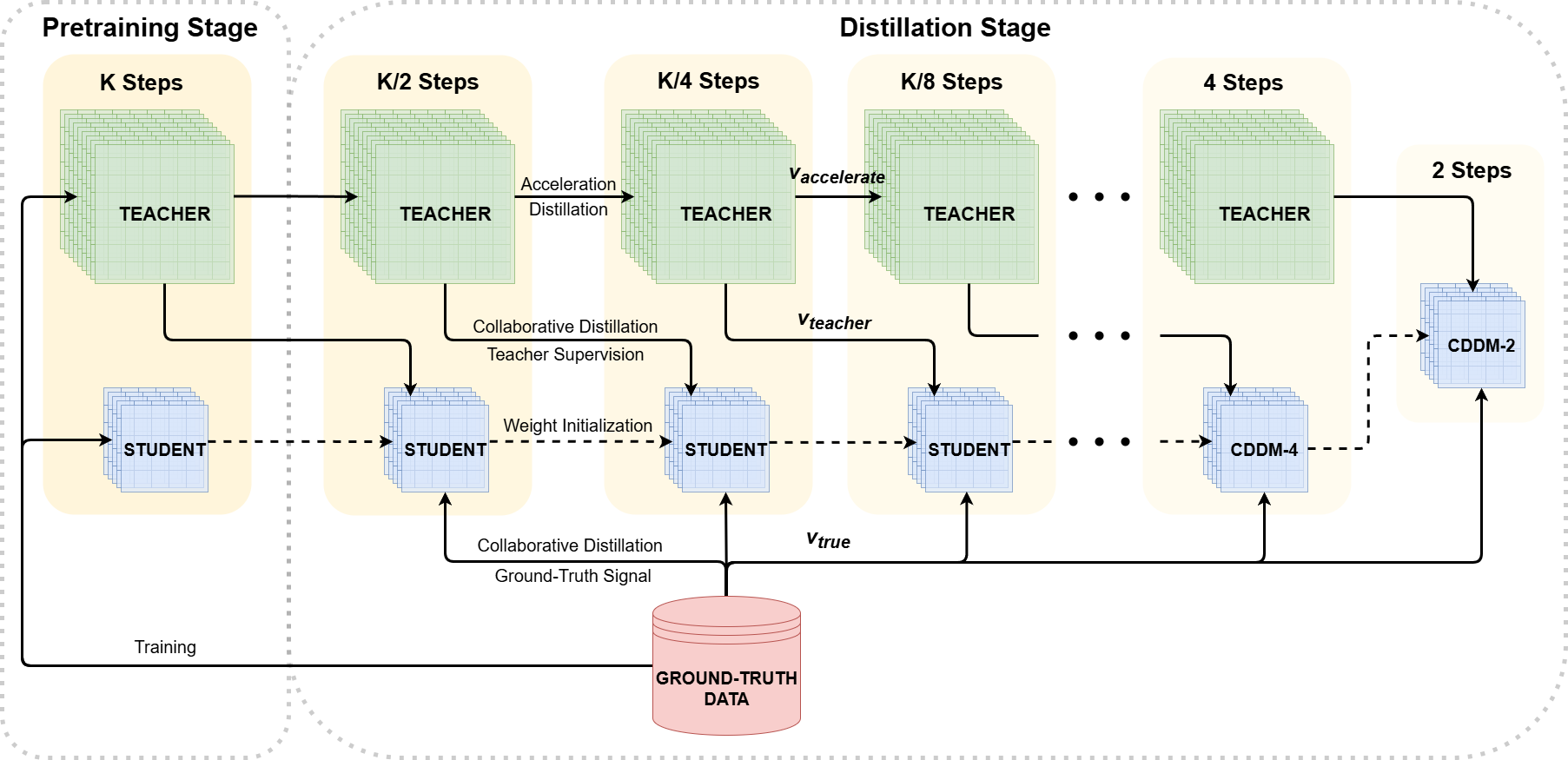}
    \caption{Framework of Collaborative-Distilled Diffusion Models}
    \label{fig:framework}
\end{figure*}

This section provides a detailed introduction to our proposed progressive collaborative knowledge distillation methodology for accelerating and compressing diffusion models in trajectory prediction. First, we briefly describe the architecture of the trajectory prediction network. Next, we present the novel Collaborative Progressive Distillation framework, illustrated with algorithmic details. In this framework, a two-stage collaborative distillation process is designed to progressively accelerate compressed models while ensuring stable and accurate training, as shown in Figure \ref{fig:framework}. Finally, we introduce an innovative training loss that distills combined knowledge from both the original data and the teacher model, effectively enhancing knowledge transfer and regularizing training to prevent overfitting.

\subsection{Trajectory Prediction Network Architecture}
\label{sec:network}
Our trajectory prediction model adopts an encoder-decoder architecture for historical contextual feature extraction and future trajectory generation. The encoder network $\mathcal{F}_\varphi$ consists of a GNN and LSTM. It processes the observed historical states of the ego agent (position, velocity, acceleration, heading), together with its social interactions, to generate a contextual embedding $\bm{f}$ as output. The decoder network $\mathcal{F}_\theta$ performs as the diffusion backbone to iteratively denoise the noisy future trajectory, conditioned on the diffusion steps and contextual embedding. Our distillation framework is model-agnostic, and we adopt the architecture from Motion Indeterminacy Diffusion~\cite{mid} to conduct our investigation of diffusion model acceleration and compression.

Our distillation process is conducted on the decoder diffusion model, as it has a significantly higher computational burden compared to the encoder. A detailed comparison can be found in the experimental study. Following MID, the decoder is a Transformer-based architecture designed to model Gaussian transitions in the diffusion process. Specifically, the decoder network is formulated as:
\begin{equation}
    \mathcal{F}_\theta(\bm{y}_k, \beta_k, \bm{f}) = 
\mathrm{MLP}\left( 
    \mathrm{Transformer}\left( 
        \mathrm{Embed}(\bm{y}_k, \beta_k, \bm{f}) 
    \right) 
\right)
\end{equation}
where $\bm{y}_k$ is noised trajectory, $\beta_k$ is timestep embedding, and $\bm{f}$ is context features from the encoder. These inputs are upsampled and fused via fully connected layers $\mathrm{Embed}(\cdot)$, enriched with sinusoidal positional embeddings to capture temporal information, and then passed through a multi-layer Transformer denoted as $\mathrm{Transformer}(\cdot)$ to model complex spatial-temporal dependencies. A final fully connected layer $\mathrm{MLP}(\cdot)$ downsamples output to project back to trajectory space.

For model compression, we uniformly reduce the hidden dimension throughout the decoder network, while preserving the overall architectural framework. This approach maintains the model structure and spatial-temporal reasoning capacity, enabling efficient model size reduction with lower computational cost.

\subsection{Collaborative Progressive Distillation Framework}

\begin{algorithm}[t]
\caption{Collaborative Progressive Distillation}
\label{alg:cpd}
\begin{algorithmic}[1]
\Require Teacher network $\mathcal{F}_\eta$; Student network $\mathcal{F}_\theta$; 

Context encoder $\mathcal{F}_\varphi$; Dataset $\mathcal{D}$; 

Student sampling steps $K$; Noise schedule $\alpha_k, \beta_k, \sigma_k$

\State \textbf{// Progressive distillation iterations}
\For{$n = 1, 2, \dots, N$}
    \If{$n = 1$}
        \State $\eta \gets \eta_{\text{pretrained}}$
        \State $\theta \gets \theta_{\text{pretrained}}$
        \Comment{Initialize from pretrained weights}
    \Else
        \State $\eta \gets \eta^{(n-1)}$
        \State $\theta \gets \theta^{(n-1)}$
        \Comment{Initialize from previous iteration}
    \EndIf

    \While{not converged}
        \State Sample data $\bm{y}_0 \sim \mathcal{D}$, noise $\boldsymbol{\epsilon} \sim \mathcal{N}(\bm{0}, \bm{I})$
        \State Sample timestep $i \sim \mathcal{U}\{1, \dots, K\}$ and set $k = i / K$
        \State Encode context: $\bm{f} = \mathcal{F}_\varphi(\bm{X}, \bm{C})$
        \State Generate noisy input: $\bm{y}_k = \alpha_k \bm{y}_0 + \sigma_k \boldsymbol{\epsilon}$

        \State \textbf{// Two-step DDIM with large teacher}
        \State $k' = k - 0.5 / K,\quad k'' = k - 1 / K$
        \State $\hat{\bm{v}}_{k'} = \mathcal{F}_\eta(\bm{y}_k, \beta_k, \bm{f})$
        \State Estimate $\hat{\bm{y}}_0 = \alpha_k \bm{y}_k - \sigma_k \hat{\bm{v}}_{k'}$
        \State $\bm{y}_{k'} = \alpha_{k'} \hat{\bm{y}}_0 + \frac{\sigma_{k'}}{\sigma_k} (\bm{y}_k - \alpha_k \hat{\bm{y}}_0)$
        \State $\hat{\bm{v}}_{k''} = \mathcal{F}_\eta(\bm{y}_{k'}, \beta_{k'}, \bm{f})$
        \State Estimate $\hat{\bm{y}}_0' = \alpha_{k'} \bm{y}_{k'} - \sigma_{k'} \hat{\bm{v}}_{k''}$
        \State $\hat{\boldsymbol{\epsilon}}_{k''} = \frac{\bm{y}_k - \alpha_{k''} \hat{\bm{y}}_0'}{\sigma_{k''}}$

        \State Construct velocity targets:
        \State $\bm{v}_{\text{teacher}} = \alpha_{k''} \hat{\boldsymbol{\epsilon}}_{k''} - \sigma_{k''} \hat{\bm{y}}_0'$
        \State $\bm{v}_{\text{true}} = \alpha_{k''} \boldsymbol{\epsilon} - \sigma_{k''} \bm{y}_0$

        \State \textbf{// One-step DDIM with small student}
        \State $\hat{\bm{v}}_{\text{student}} = \mathcal{F}_\theta(\bm{y}_k, \beta_{k''}, \bm{f})$
        \State Compute regularized loss:
        \State $\mathcal{L}_\theta = (1-\lambda) \| \hat{\bm{v}}_{\text{student}} - \bm{v}_{\text{teacher}} \|^2 + \lambda \| \hat{\bm{v}}_{\text{student}} - \bm{v}_{\text{true}} \|^2$
        \State Update student: $\theta \gets \theta - \gamma \nabla_\theta \mathcal{L}_\theta$

        \State \textbf{// Collaboratively distill next teacher}
        \State $\hat{\bm{v}}_{\text{accelerate}} = \mathcal{F}_\eta(\bm{y}_k, \beta_{k''}, \bm{f})$
        \State $\mathcal{L}_\eta = \| \hat{\bm{v}}_{\text{accelerate}} - \bm{v}_{\text{teacher}} \|^2$
        \State Update teacher: $\eta \gets \eta - \gamma \nabla_\eta \mathcal{L}_\eta$
    \EndWhile

    \State $\eta^{(n)} \gets \eta$ \Comment{Distilled teacher becomes next teacher}
    \State $\theta^{(n)} \gets \theta$ \Comment{Distilled student initializes next student}
    \State $K \gets K / 2$ \Comment{Halve the number of sampling steps}
\EndFor
\end{algorithmic}
\end{algorithm}

While diffusion models have shown superior accuracy in trajectory prediction, their multi-step denoising process and relatively large backbone size impose significant computational burdens and extended inference times, limiting their applicability in real-world trajectory prediction particularly on edge computing devices. Although prior studies have investigated efficient diffusion models, as discussed in Section~\ref{sec:efficient}, existing approaches typically address either acceleration or compression individually. None of research has effectively achieved both objectives simultaneously.

\subsubsection{Framework Overview}
To tackle above-mentioned obstacles, we propose Collaborative Progressive Distillation (CPD) framework to both accelerate and compress diffusion models iteratively for efficient trajectory prediction. We begin by initializing a large teacher model and a lightweight student model by training both models for $K$ steps on the data. Subsequently, the student model iteratively distills from the teacher model to reduce its sampling steps by half. In each iteration, we also distill the teacher model to $K/2$ steps to server as the teacher for the next iteration, and the student model further shortens its steps based on the distilled teacher model. This collaborative distillation framework enables the effective and progressive transfer of knowledge from the original large, slow, but highly accurate diffusion model to a compact and fast student model, continuing until the number of sampling steps reaches the desired target. The detailed CPD algorithm is presented in Algorithm~\ref{alg:cpd}. We adopt DDIM as the sampling parameterization method in our framework.

\subsubsection{Pretraining Stage}
In the pretraining stage, both a large and a small diffusion model, $\eta_{\text{pretrained}}$ and $\theta_{\text{pretrained}}$, are independently trained on the original dataset $\mathcal{D}$ using the standard DDPM training approach~\cite{ddpm}. Unlike the original formulation, the training target is set to velocity-space prediction \cite{progressive} rather than noise prediction, which facilitates more stable distillation in the subsequent stage. Pretraining allows the large diffusion model to effectively capture the original data distribution and provides the small model with an informative initialization, enabling faster convergence. 

\subsubsection{Distillation Stage}
Different from Progressive Distillation~\cite{progressive} that cannot directly copy weights between the teacher and student for initialization due to different model sizes, we iteratively and collaboratively distill a large model to serve as the teacher for distillation supervision and a small model to initialize the student. Moreover, directly applying a pretrained small model may fail to capture the original data distribution due to its limited representational capacity. However, our method only uses it to provide a warm-start initialization for training and is subsequently distilled under the guidance of a more knowledgeable large model. This approach improves the stability and efficiency of the training convergence process.

In the first iteration, the student model $\mathcal{F}_\theta$ is initialized by copying the pretrained small model, $\theta \gets \theta_{\text{pretrained}}$, and its number of sampling steps is reduced to half, $K \gets K/2$. The teacher model $\mathcal{F}_\eta$ remains the original pretrained large model. The context encoder $\mathcal{F}_\varphi$ remains frozen throughout distillation due to its compact size. Distillation process encourages the student prediction with a single DDIM step, $\hat{\bm{v}}_{\text{student}}$, to match the target value derived from two successive DDIM steps of the teacher model, $\bm{v}_{\text{teacher}}$, thereby achieving acceleration while preserving accurate knowledge from the large pretrained model. Assuming the current timestep is $k$ and the student has $K$ sampling steps, a single student step progresses to $k - \frac{1}{K}$. This corresponds to two teacher steps at $k - \frac{0.5}{K}$ and $k - \frac{1}{K}$, since the teacher employs twice as many sampling steps as the student. This setting of continuous sampling steps mentioned in \ref{sec:ddpm} ensures that the noise schedule of the student model is evenly aligned with the teacher model, enabling smooth and consistent knowledge transfer. In parallel, to obtain the teacher model for the next distillation iteration, a large model with $K/2$ sampling steps is distilled using the Progressive Distillation approach.

In the following iterations, the student model is collaboratively distilled by initializing its parameters from the distilled small student of the previous iteration, $\theta \gets \theta^{(n-1)}$, and using the acceleration-distilled large teacher from the previous iteration as supervision, $\eta \gets \eta^{(n-1)}$. This hierarchical strategy progressively reduces the number of sampling steps following the sequence $\{K, \frac{K}{2},  \frac{K}{4}, \dots, 2, 1\}$, enabling effective knowledge transfer across models of different capacities and sampling steps. As such, the CPD framework simultaneously compresses and accelerates a large, well-trained diffusion model into a lightweight model with significantly fewer sampling steps, while preserving accuracy through an efficient knowledge flow.

\subsection{Collaborative Progressive Distillation Training Loss}
\label{sec:loss}

This section details the construction of the Collaborative Progressive Distillation training loss, which extends velocity-prediction objective as described in Section~\ref{sec:v-pre}, to enable robust, stable, and effective knowledge distillation for diffusion model compression and acceleration.

\subsubsection{Student Distillation Loss}
The training loss for distilling a small student diffusion model is defined as:
\begin{equation}
    \mathcal{L}_{\text{student}} = (1-\lambda) \| \hat{\bm{v}}_{\text{student}} - \bm{v}_{\text{teacher}} \|^2 + \lambda \| \hat{\bm{v}}_{\text{student}} - \bm{v}_{\text{true}} \|^2
\end{equation}
where $\hat{\bm{v}}_{\text{student}}$ is obtained via a single DDIM step using the small student model, $\bm{v}_{\text{teacher}}$ is computed through two DDIM steps with the large teacher model, and $\lambda$ is a tunable regularization hyperparameter. Note that the teacher remains frozen during the training process and only the student parameters are updated. This loss design enables simultaneous acceleration and compression by distilling knowledge from the slow large model into the fast lightweight model within a single unified training step. Additionally, we empirically observe that the student model tends to overfit to the teacher during distillation. Moreover, if the teacher model is not well-trained, the student may inherit its biases or suboptimal performance. To mitigate this, we introduce a regularization term into the distillation loss that encourages the student prediction to remain close to the ground truth signal $\bm{v}_{\text{true}}$ derived from the original data $\bm{y}_0$ and the sampled random noise $\boldsymbol{\epsilon}$:
\begin{equation}
    \bm{v}_{\text{true}} = \alpha_{k''} \boldsymbol{\epsilon} - \sigma_{k''} \bm{y}_0
\end{equation}
where $k''=k-\frac{1}{K}$ denotes one forward step of the student model, given the current timestep $k$ and student sampling steps $K$, in order to align the timestep with the student’s prediction for a consistent training objective. Experimental results demonstrate that the proposed regularization effectively stabilizes the distillation process and mitigates overfitting, while preserving model accuracy through knowledge transfer from both the trained teacher and the ground-truth data.

\subsubsection{Teacher Distillation Loss}
In each distillation iteration, an accelerated large teacher model with half the sampling steps is distilled in parallel to serve as the intermediate teacher for the next iteration. This design ensures greater model capacity for more accurate supervisory knowledge transfer. The corresponding training loss is defined as:
\begin{equation}
    \mathcal{L}_{\text{accelerate}} = \| \hat{\bm{v}}_{\text{accelerate}} - \bm{v}_{\text{teacher}} \|^2
\end{equation}
where $\hat{\bm{v}}_{\text{accelerate}}$ is obtained via a single DDIM step using an accelerated large model, and $\bm{v}_{\text{teacher}}$ is obtained using two DDIM steps of the teacher model as described previously. The initialization of the accelerated model is performed by copying the teacher model, following the same approach as in Progressive Distillation~\cite{progressive}.

\section{Experiments}
\label{sec:experiment}
This section demonstrates state-of-the-art experimental performance of our proposed CDDM for diffusion model acceleration and compression on widely-used public trajectory prediction benchmarks. Ablation study shows the effectiveness of our framework compared to alternative implementation methods for diffusion model distillation. Additionally, qualitative visualization results display CDDM’s ability to generate diverse and accurate trajectory predictions.

\subsection{Experimental Setup}
\subsubsection{Dataset}
CDDM is evaluated on two widely used public trajectory prediction benchmarks: the ETH-UCY dataset for pedestrians and the nuScenes dataset for vehicles. The ETH-UCY dataset records pedestrian movements at 2.5 Hz, capturing dynamic social interactions and naturalistic crowd behaviors across 5 diverse scenarios. The nuScenes dataset consists of 1,000 urban driving scenes, where the ego vehicle and surrounding agents are annotated and consistently tracked at 2 Hz. In both datasets, positions are provided in world coordinates, and thus all evaluation results are reported in meters.

\subsubsection{Evaluation Metrics}
The standard evaluation metrics Average Displacement Error (ADE) and Final Displacement Error (FDE) are used to evaluate trajectory prediction performance. ADE computes the mean Euclidean distance between the predicted and ground truth trajectories over all timesteps, while FDE only measures the final position. Due to the probabilistic nature of our trajectory prediction method, we adopt a best-of-N evaluation strategy, where the prediction with the lowest error is selected from $N=20$ stochastic samples. The pedestrian trajectory prediction takes 3.2 seconds of historical trajectory (8 timesteps) as observation input, and 4.8 seconds of future trajectory (12 timesteps) as prediction target. The vehicle trajectory prediction uses 4 seconds of historical trajectory (8 timesteps) to predict 3 seconds of future trajectory (6 timesteps). We also report the number of parameters as a measure of model size, floating-point operations (FLOPs) as computational cost, and prediction inference time as latency.

\begin{table}[!h]
\centering
\renewcommand{\arraystretch}{1.2}
\caption{Model complexity comparison between baseline and lightweight variants.}
\label{tab:model_size_comparison}
\resizebox{\columnwidth}{!}{
\begin{tabular}{l|cc|cc}
\hline\hline
\multirow{2}{*}{\textbf{Component}} &
\multicolumn{2}{c|}{\textbf{Baseline} ($H=256$)} &
\multicolumn{2}{c}{\textbf{Lightweight} ($H=16$)} \\
\cline{2-5}
& \small Params & \small FLOPs & \small Params & \small FLOPs\\
\hline
Encoder & 175K & 1.19M & 175K & 1.19M \\
Decoder & 9.04M & 50.25M & 56K & 0.64M\\
\hline
\textbf{Total} & \textbf{9.22M} & \textbf{51.44M} &  \textbf{231K} & \textbf{1.83M}\\
\hline\hline
\end{tabular}
}
\end{table}

\subsubsection{Implementation Details}

We employ a Transformer-based architecture for diffusion backbone model as introduced in \ref{sec:network}. The model processes input sequences of 2D points \( \bm{x} \in \mathbb{R}^{B \times P \times 2} \), where \( B \) is the batch size and \( P \) is the prediction horizon. A scalar variance schedule \( \beta \in \mathbb{R}^{B} \) is encoded using sinusoidal functions and concatenated with context features \( \bm{f} \in \mathbb{R}^{B \times F} \) from the encoder network to form a context embedding used to condition all subsequent layers. The input \( \bm{x} \) is projected to \( \mathbb{R}^{B \times P \times 2H} \) using a context-conditioned linear layer. A fixed sinusoidal positional encoding is added, and the sequence is then passed through a Transformer encoder composed of 3 layers, each with 4 attention heads and feedforward dimension \( 2H \). The output dimension is further reduced through three context-conditioned linear layers following $2H \rightarrow H \rightarrow H/2 \rightarrow 2$. The encoder dimension $F$ is fixed at 256. The hidden dimension $H$ is reduced to compress the decoder diffusion model while maintaining the network structure. Since the decoder is significantly larger than the encoder in the baseline implementation, as shown in Table~\ref{tab:model_size_comparison}, we compress and distill only the diffusion backbone in the decoder and keep the encoder frozen. The model takes position, velocity, acceleration, and heading as input and predicts future velocity, which is then integrated over time using a single-integrator dynamics model to obtain the final trajectory. All inputs to the diffusion model are standardized before feeding into the network. During training, the model is optimized using AdamW with an initial learning rate of 0.001, which decays linearly to zero. The Exponential Moving Average (EMA) technique is adopted to stabilize distillation. All the experiments are conducted on a single NVIDIA RTX A6000 GPU.


\definecolor{LighterCyan}{rgb}{0.93,1,1}
\definecolor{LightCyan}{rgb}{0.85,1,1}

\begin{table*}[t]
\centering
\caption{Performance comparison of $minADE_{20}$ and $minFDE_{20}$ (meters) between CDDM, DDIM, and PD across datasets. Bold fonts represent the best result within the same model size. Lower is better.}
\label{tab:performance_comparison}
\renewcommand{\arraystretch}{1.2}
\setlength{\tabcolsep}{4pt}
\begin{tabular}{c|c|c|c|c>{\columncolor[gray]{0.95}}cc>{\columncolor[gray]{0.95}}cc>{\columncolor[gray]{0.95}}cc>{\columncolor[gray]{0.95}}cc>{\columncolor[gray]{0.95}}c>{\columncolor{LighterCyan}{}}c>{\columncolor{LightCyan}{}}c>{\columncolor{LighterCyan}{}}c>{\columncolor{LightCyan}{}}c}
\hline\hline
\multirow{3}{*}{\textbf{Type}} & \multirow{3}{*}{\textbf{Dataset}} & \multirow{3}{*}{\textbf{Method}} & \multirow{3}{*}{\shortstack{\textbf{Backbone} \\ \textbf{Parameters}}} & \multicolumn{14}{c}{\textbf{Sampling Steps}} \\
\cline{5-18}
 & & & & \multicolumn{2}{c}{\textbf{128}} & \multicolumn{2}{c}{\textbf{64}} & \multicolumn{2}{c}{\textbf{32}} & \multicolumn{2}{c}{\textbf{16}} & \multicolumn{2}{c}{\textbf{8}} & \multicolumn{2}{c}{\textbf{4}} & \multicolumn{2}{c}{\textbf{2}} \\
\cline{5-18}
 & & & & \textbf{ADE} & \textbf{FDE} & \textbf{ADE} & \textbf{FDE} & \textbf{ADE} & \textbf{FDE} & \textbf{ADE} & \textbf{FDE} & \textbf{ADE} & \textbf{FDE} & \textbf{ADE} & \textbf{FDE} & \textbf{ADE} & \textbf{FDE} \\
\hline
\multirow{24}{*}{\textbf{Pedestrian}}
& \multirow{4}{*}{\textbf{ETH}}
& Baseline & 9,043K
& 0.41 & 0.70 & 0.42 & 0.70 & 0.42 & 0.69 & 0.43 & 0.71 & 0.45 & 0.69 & 0.46 & 0.69 & 0.49 & 0.68 \\
\cline{3-18}
& & DDIM  & \multirow{3}{*}{56K}
& 0.43 & 0.76 & 0.43 & 0.75 & 0.43 & 0.75 & 0.42 & 0.74 & 0.44 & 0.80 & \textbf{0.46} & 0.84 & 0.52 & 1.02 \\
& & PD
& 
& 0.43 & 0.76 & 0.47 & 0.74 & 0.44 & 0.73 & 0.45 & 0.78 & 0.47 & 0.77 & 0.49 & 0.78 & 0.58 & 0.77 \\
& & \textbf{CDDM}
& 
& 0.43 & 0.76 & 0.43 & 0.74 & 0.4
3 & 0.69 & 0.44 & 0.69 & 0.44 & 0.70 & \textbf{0.46} & \textbf{0.69} & \textbf{0.49} & \textbf{0.74} \\
\cline{2-18}
& \multirow{4}{*}{\textbf{Hotel}}
& Baseline & 9,043K
& 0.17 & 0.28 & 0.16 & 0.28 & 0.17 & 0.30 & 0.17 & 0.31 & 0.19 & 0.33 & 0.18 & 0.29 & 0.18 & 0.31 \\
\cline{3-18}
& & DDIM  & \multirow{3}{*}{56K}
& 0.17 & 0.29 & 0.17 & 0.29 & 0.16 & 0.28 & 0.17 & 0.29 & 0.18 & 0.30 & 0.19 & 0.32 & 0.24 & 0.43 \\
& & PD    &
& 0.17 & 0.29 & 0.19 & 0.33 & 0.20 & 0.35 & 0.20 & 0.36 & 0.21 & 0.36 & 0.22 & 0.38 & 0.22 & 0.36 \\
& & \textbf{CDDM}  &
& 0.17 & 0.29 & 0.17 & 0.29 & 0.19 & 0.32 & 0.19 & 0.33 & 0.17 & 0.28 & \textbf{0.18} & \textbf{0.31} & \textbf{0.18} & \textbf{0.31} \\
\cline{2-18}
& \multirow{4}{*}{\textbf{Univ}}
& Baseline & 9,043K
& 0.22 & 0.42 & 0.22 & 0.42 & 0.23 & 0.42 & 0.23 & 0.42 & 0.24 & 0.43 & 0.24 & 0.43 & 0.25 & 0.43 \\
\cline{3-18}
& & DDIM  & \multirow{3}{*}{56K}
& 0.23 & 0.45 & 0.23 & 0.45 & 0.23 & 0.45 & 0.23 & 0.46 & 0.24 & 0.48 & \textbf{0.26} & 0.51 & 0.30 & 0.62 \\
& & PD    &
& 0.23 & 0.45 & 0.24 & 0.47 & 0.25 & 0.46 & 0.25 & 0.43 & 0.25 & 0.44 & 0.26 & 0.45 & 0.29 & 0.45 \\
& & \textbf{CDDM}  &
& 0.23 & 0.45 & 0.23 & 0.42 & 0.24 & 0.42 & 0.23 & 0.43 & 0.24 & 0.42 & \textbf{0.26} & \textbf{0.44} & \textbf{0.27} & \textbf{0.44} \\
\cline{2-18}
& \multirow{4}{*}{\textbf{Zara1}}
& Baseline & 9,043K
& 0.21 & 0.38 & 0.20 & 0.37 & 0.21 & 0.40 & 0.21 & 0.40 & 0.22 & 0.42 & 0.21 & 0.38 & 0.21 & 0.39 \\
\cline{3-18}
& & DDIM  & \multirow{3}{*}{56K}
& 0.25 & 0.45 & 0.24 & 0.44 & 0.24 & 0.44 & 0.24 & 0.44 & 0.24 & 0.46 & 0.24 & 0.46 & 0.27 & 0.55 \\
& & PD    &
& 0.25 & 0.45 & 0.25 & 0.48 & 0.27 & 0.54 & 0.28 & 0.57 & 0.31 & 0.62 & 0.32 & 0.63 & 0.32 & 0.62 \\
& & \textbf{CDDM}  &
& 0.25 & 0.45 & 0.21 & 0.41 & 0.21 & 0.41 & 0.21 & 0.41 & 0.21 & 0.39 & \textbf{0.21} & \textbf{0.40} & \textbf{0.22} & \textbf{0.40} \\
\cline{2-18}
& \multirow{4}{*}{\textbf{Zara2}}
& Baseline & 9,043K
& 0.17 & 0.32 & 0.16 & 0.32 & 0.17 & 0.34 & 0.17 & 0.35 & 0.18 & 0.35 & 0.18 & 0.33 & 0.18 & 0.36 \\
\cline{3-18}
& & DDIM  & \multirow{3}{*}{56K}
& 0.19 & 0.35 & 0.19 & 0.34 & 0.19 & 0.35 & 0.19 & 0.34 & 0.19 & 0.34 & \textbf{0.19} & \textbf{0.35} & \textbf{0.20} & 0.38 \\
& & PD    &
& 0.19 & 0.35 & 0.19 & 0.36 & 0.20 & 0.38 & 0.21 & 0.40 & 0.22 & 0.42 & 0.24 & 0.43 & 0.23 & 0.46 \\
& & \textbf{CDDM}  &
& 0.19 & 0.35 & 0.17 & 0.34 & 0.20 & 0.36 & 0.21 & 0.38 & 0.18 & 0.34 & \textbf{0.19} & 0.36 & \textbf{0.20} & \textbf{0.37} \\
\cline{2-18}
& \multirow{4}{*}{\textbf{AVG}}
& Baseline & 9,043K
& 0.24 & 0.42 & 0.23 & 0.42 & 0.24 & 0.43 & 0.24 & 0.44 & 0.25 & 0.45 & 0.25 & 0.42 & 0.26 & 0.43 \\
\cline{3-18}
& & DDIM  & \multirow{3}{*}{56K}
& 0.25 & 0.46 & 0.25 & 0.45 & 0.25 & 0.45 & 0.25 & 0.46 & 0.26 & 0.47 & 0.27 & 0.50 & 0.31 & 0.60 \\
& & PD    &
& 0.25 & 0.46 & 0.27 & 0.48 & 0.27 & 0.49 & 0.28 & 0.51 & 0.29 & 0.52 & 0.30 & 0.53 & 0.33 & 0.53 \\
& & \textbf{CDDM}  &
& 0.25 & 0.46 & 0.24 & 0.44 & 0.25 & 0.44 & 0.26 & 0.45 & 0.25 & 0.43 & \textbf{0.26} & \textbf{0.44} & \textbf{0.27} & \textbf{0.45} \\
\hline
\multirow{4}{*}{\textbf{Vehicle}}
& \multirow{4}{*}{\textbf{nuScenes}}
& Baseline & 9,043K
& 0.47 & 0.76 & 0.52 & 0.86 & 0.54 & 0.88 & 0.55 & 0.87 & 0.54 & 0.93 & 0.58 & 0.97 & 0.66 & 1.05 \\
\cline{3-18}
& & DDIM  & \multirow{3}{*}{56K}
& 0.51 & 0.87 & 0.51 & 0.83 & 0.51 & 0.84 & 0.51 & 0.85 & 0.55 & 0.92 & 0.61 & 1.00 & \textbf{0.68} & 1.07 \\
& & PD    &
& 0.51 & 0.87 & 0.54 & 0.93 & 0.57 & 0.93 & 0.56 & 0.94 & 0.59 & 0.97 & 0.62 & 1.09 & 0.74 & 1.02 \\
& & \textbf{CDDM}  &
& 0.51 & 0.87 & 0.52 & 0.85 & 0.52 & 0.89 & 0.54 & 0.88 & 0.54 & 0.88 & \textbf{0.59} & \textbf{0.95} & 0.69 & \textbf{0.96} \\
\hline\hline
\end{tabular}
\end{table*}

\subsection{Performance Analysis}

To evaluate the performance of our proposed method, we compare our Collaborative-Distilled Diffusion Models (CDDM) trained using the Collaborative Progressive Distillation (CPD) framework, against state-of-the-art diffusion model acceleration approaches, i.e., Denoising Diffusion Implicit Models (DDIM) and Progressive Distillation (PD). We first train a 128-step diffusion model using the standard $v$-parameterized DDPM training approach with hidden dimension $H=256$, which serves as the baseline. PD is then applied to accelerate the baseline model from 128 down to 2 sampling steps while maintaining its capacity. Next, our CDDM is distilled from the baseline model using a compressed lightweight model with $H=16$, which contains approximately $161\times$ fewer parameters in the diffusion backbone and $79\times$ fewer FLOPs per sampling step compared to the baseline (as shown in Table \ref{tab:model_size_comparison}). For fair comparison, DDIM and PD, along with our CPD, are also applied to progressively accelerate the lightweight models from 128 to 2 sampling steps by halving the number of steps in each iteration. In our CPD framework, the lightweight student model is distilled from the baseline model as the teacher. 

\begin{table*}[!t]
\centering
\caption{Computational cost and $minADE_{20}/minFDE_{20}$ (meters) comparison of CDDM against state-of-the-art trajectory prediction methods on ETH-UCY. Lower is better.}
\label{tab:sota}
\setlength{\tabcolsep}{4pt}
\renewcommand{\arraystretch}{1.15}
\resizebox{\textwidth}{!}{
\begin{tabular}{c|c|c|c|c>{\columncolor[gray]{0.95}}cc>{\columncolor[gray]{0.95}}cc>{\columncolor[gray]{0.95}}cc>{\columncolor[gray]{0.95}}cc>{\columncolor[gray]{0.95}}c>{\columncolor{LighterCyan}{}}c>{\columncolor{LightCyan}{}}c}
\hline \hline
\multirow{2}{*}{\textbf{Method}} & 
\multirow{2}{*}{\shortstack{\textbf{Model} \\ \textbf{Parameters}}} & 
\multirow{2}{*}{\textbf{FLOPs}} & 
\multirow{2}{*}{\textbf{Inference}} & \multicolumn{2}{c}{\textbf{ETH}}  & 
\multicolumn{2}{c}{\textbf{Hotel}} & 
\multicolumn{2}{c}{\textbf{Univ}} & 
\multicolumn{2}{c}{\textbf{Zara1}} & 
\multicolumn{2}{c}{\textbf{Zara2}}  & 
\multicolumn{2}{c}{\textbf{AVG}} \\
\cline{5-16}
&  &  &  & \textbf{ADE} & \textbf{FDE} & \textbf{ADE} & \textbf{FDE} & \textbf{ADE} & \textbf{FDE} & \textbf{ADE} & \textbf{FDE} & \textbf{ADE} & \textbf{FDE} & \textbf{ADE} & \textbf{FDE} \\
\hline
Social-GAN (\textcolor{blue}{CVPR'18}) \cite{social_gan}     & 46.4K & 3.16M & $\sim$11 ms & 0.81  & 1.52 & 0.72  & 1.61 & 0.60  & 1.26 & 0.34  & 0.69 & 0.42  & 0.84 & 0.58  & 1.18 \\
STAR (\textcolor{blue}{ECCV'20}) \cite{STAR}          & 965K & 16.88M & $\sim$31 ms & 0.36  & 0.65 & 0.17  & 0.36 & 0.31  & 0.62 & 0.29  & 0.52 & 0.22  & 0.46 & 0.26  & 0.53 \\
Social-STGCNN (\textcolor{blue}{CVPR'20}) \cite{social_stgcnn}        & 7.6K & 1.96M & $\sim$2 ms & 0.64  & 1.11 & 0.49  & 0.85 & 0.44  & 0.79 & 0.34  & 0.53 & 0.30 & 0.48 & 0.44  & 0.75 \\
Trajectron++ (\textcolor{blue}{ECCV'20}) \cite{trajectron++}   & 10.13M & 143.60M & $\sim$29 ms & 0.61  & 1.02 & 0.19  & 0.28 & 0.30  & 0.54 & 0.24  & 0.42 & 0.18  & 0.32 & 0.30  & 0.51 \\
Agentformer (\textcolor{blue}{ICCV'21}) \cite{agentformer}    & 6.78M & 3.08G & $\sim$52 ms & 0.45  & 0.75 & 0.14  & 0.22 & 0.25  & 0.45 & 0.18  & 0.30 & 0.14  & 0.24 & 0.23  & 0.39 \\
MemoNet (\textcolor{blue}{CVPR'22}) \cite{memonet}         & 5.28M & 408.35M & $\sim$21 ms & 0.40  & 0.61 & 0.11  & 0.17 & 0.24  & 0.43 & 0.18  & 0.32 & 0.14  & 0.24 & 0.21  & 0.35 \\
GroupNet (\textcolor{blue}{CVPR'22}) \cite{groupnet}       & 3.14M & 221.06M & $\sim$16 ms & 0.46  & 0.73 & 0.15  & 0.25 & 0.26  & 0.49 & 0.21  & 0.39 & 0.17  & 0.33 & 0.25  & 0.44 \\
MID (\textcolor{blue}{CVPR'22}) \cite{mid}   & 9.22M & 5.03G & $\sim$223 ms & 0.39  & 0.66 & 0.13  & 0.22 & 0.22  & 0.45 & 0.17  & 0.30 & 0.13  & 0.27 & 0.21  & 0.38 \\
LED (\textcolor{blue}{CVPR'23}) \cite{leapfrog}    & 11.2M & 41.46M & $\sim$46 ms & 0.39  & 0.58 & 0.11  & 0.17 & 0.26  & 0.43 & 0.18  & 0.26 & 0.13  & 0.22 & 0.21  & 0.33 \\
EqMotion (\textcolor{blue}{CVPR'23}) \cite{eqmotion}         & 3.02M & 7.75M & $\sim$12 ms & 0.40  & 0.61 & 0.12  & 0.18 & 0.23  & 0.43 & 0.18  & 0.32 & 0.13  & 0.23 & 0.21  & 0.35 \\
MRGTraj (\textcolor{blue}{TCSVT'24}) \cite{mrgtraj}         & 4.36M & 580.38M & $\sim$27 ms & 0.28  & 0.47 & 0.21  & 0.39 & 0.33  & 0.60 & 0.24  & 0.44 & 0.22  & 0.41 & 0.26  & 0.46 \\
\hline
\textbf{CDDM (Ours)}      & 231K & 3.75M & $\sim$9 ms & 0.46  & 0.69 & 0.18  & 0.31 & 0.26  & 0.44 & 0.21  & 0.40 & 0.19  & 0.36 & 0.26  & 0.44 \\
\hline \hline
\end{tabular}
}
\end{table*}

\begin{table*}[!t]
\centering
\caption{Comparison across hidden dimensions of CDDM with $K=4$ and $K=2$ on ETH.}
\label{tab:hiddendim}
\begin{tabular}{c|c|c|cccc|cccc}
\hline \hline
\textbf{Hidden} & \textbf{Encoder} & \textbf{Decoder} &
\multicolumn{4}{c|}{$K=4$} & \multicolumn{4}{c}{$K=2$}  \\ 
\cline{4-11}
\textbf{Dimension} & \textbf{Parameters} & \textbf{Parameters} & \textbf{FLOPs} & \textbf{Inference} & \textbf{ADE} & \textbf{FDE} & \textbf{FLOPs} & \textbf{Inference} & \textbf{ADE} & \textbf{FDE}\\ 
\hline
\textbf{256} & \multirow{3}{*}{175K} & 9,043K & 202.19M & $\sim$9.8 ms & 0.46 & 0.69 & 101.69M & $\sim$6.2 ms & 0.49 & 0.68\\ 
\textbf{64}  &  & 526K   & 15.51M & $\sim$9.5 ms & 0.44 & 0.69 & 8.35M & $\sim$5.8 ms & 0.51 & 0.72 \\ 
\textbf{16}  &  & 56K    & 3.75M & $\sim$9.1 ms & 0.46 & 0.69 & 2.47M & $\sim$5.6 ms & 0.49 & 0.74 \\ \hline \hline

\end{tabular}

\end{table*}

The ADE/FDE performance comparison on ETH-UCY and nuScenes datasets with the best-of-20 strategy is presented in Table~\ref{tab:performance_comparison}. Our CDDM achieves strong performance with only 4 or 2 sampling steps, highlighted in light blue. Specifically, on the ETH-UCY dataset, it maintains \emph{96.2\%} and \emph{95.5\%} of the baseline model's ADE and FDE, while using only \emph{0.62\%} of the model size and \emph{3.2\%} of the inference time (\emph{9 ms}) at 4 sampling steps. In contrast to state-of-the-art methods such as DDIM and PD, which focus solely on accelerating inference while retaining the original model size, our compressed lightweight model achieves the competitive accuracy with significantly reduced computational cost. Moreover, our proposed CPD framework offers three key advantages:

\begin{itemize}
    \item \textbf{Independence from initial training quality:} CPD is not limited by the performance of the initial 128-step compact diffusion model. Each iteration uses the previous lightweight model only as initialization,  allowing performance to recover in later stages. This is demonstrated in our experimental results, where the first iteration shows a suboptimal result, but performance improves in subsequent iterations.
    
    \item \textbf{Collaborative knowledge transfer from high-capacity teachers:} The student learns directly from a large accelerated teacher model with higher capacity in each iteration, enabling more accurate knowledge transfer. This prevents the accumulation of errors that might occur if a poorly trained student were used as the next teacher, as is the case in Progressive Distillation. This is reflected in the results where some later iterations outperform earlier ones.
    
    \item \textbf{Stable and robust training via dual-signal supervision:} CPD is more stable and resilient to fluctuations in teacher performance. This is because the distillation loss combines supervision from both the teacher model and ground-truth data. By not relying solely on the teacher’s output, the model avoids overfitting to any bias from the teacher and continues learning from the true signal. As shown in the results, CDDM sometimes achieves even better performance than the corresponding teacher from Progressive Distillation.
\end{itemize}

Overall, our method enables effective diffusion model distillation for both acceleration and compression by more accurate knowledge transfer, outperforming widely adopted state-of-the-art approaches as demonstrated by the experimental results. Furthermore, we compare our well-trained 4-step CDDM with prior state-of-the-art trajectory prediction methods on the ETH-UCY dataset, as shown in Table~\ref{tab:sota} and Figure~\ref{fig:modelsize}. Our method delivers comparable performance while substantially reducing model size, FLOPs, and latency. This achieves an optimal balance between prediction accuracy and computational efficiency, enabling real-time and lightweight trajectory prediction with diffusion models on edge computing devices.

\subsection{Ablation Study}

\begin{figure*}[t]
    \centering
    \includegraphics[width=\linewidth]{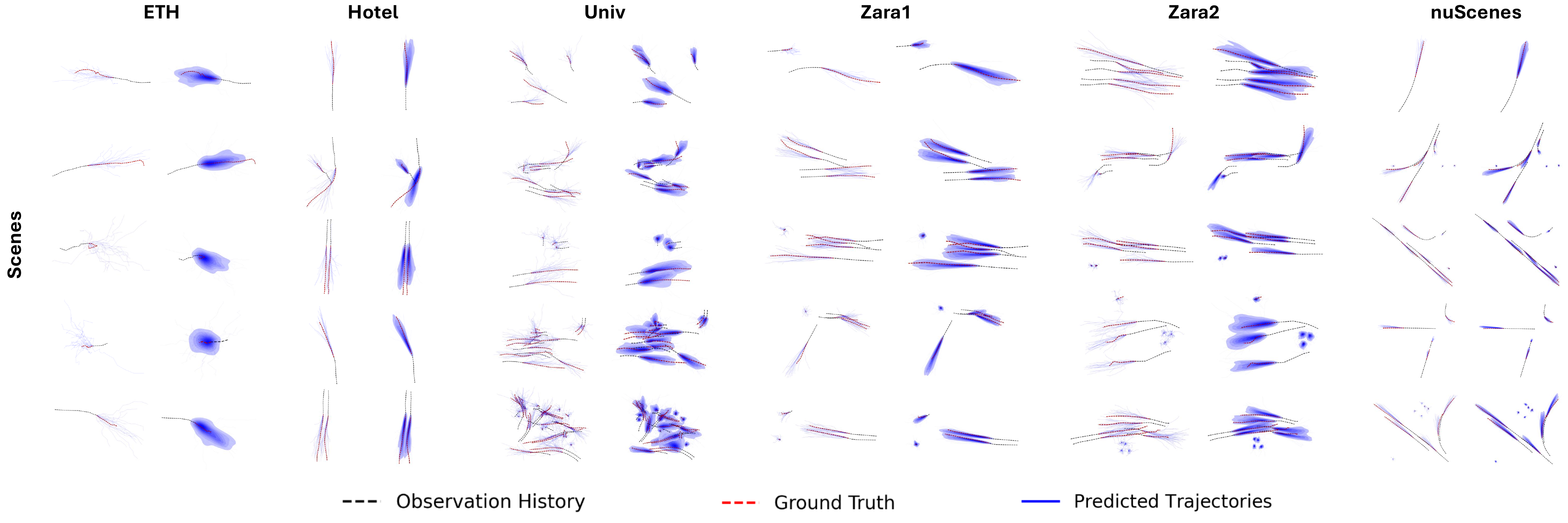}
    \caption{Visualization of CDDM trajectory predictions on the ETH-UCY (pedestrian) and nuScenes (vehicle) datasets. The left columns show 20 individually sampled trajectories, while the right columns display trajectory joint distributions generated from 50 samples.}
    \label{fig:vis}
\end{figure*}

\subsubsection{Impact of Key Components in CDDM} We conduct an ablation study to explore the effect of four key components in CDDM by evaluating its performance after removing each component individually, as shown in Table~\ref{tab:ablation}. All models in the comparison are configured with hidden dimension $H=16$ and sampling steps $K=4$, and are trained under identical settings, except that one component is ablated in each case:
\begin{itemize}
    \item \textbf{Without collaborative acceleration}: The small model is directly trained and progressively distilled without guidance from a large teacher model.
    \item \textbf{Without collaborative compression}: The small model is compressed from a large accelerated teacher at 4 sampling steps.
    \item \textbf{Without data regularization}: The distillation loss relies solely on the teacher’s signal, with the regularization weight $\lambda$ set to 0.
    \item \textbf{Without weight initialization}: The small model is distilled directly from the teacher at 4 sampling steps without parameter initialization.
\end{itemize}
The ablation results demonstrate that all the components of CDDM contribute significantly to the overall performance.

\begin{table}[h]
\centering
\caption{Ablation study on key components of CDDM on ETH.}
\label{tab:ablation}
\resizebox{\columnwidth}{!}{
\begin{tabular}{cccc|cc}
\hline \hline
\multicolumn{6}{c}{\textbf{CDDM} ($H=16$, $K=4$)}
\\ \hline
\textbf{Collaborative} & 
\textbf{Collaborative} & 
\textbf{Data} & 
\textbf{Weight} & 
\multirow{2}{*}{\textbf{ADE}} & 
\multirow{2}{*}{\textbf{FDE}} \\ 
\textbf{Acceleration} & 
\textbf{Compression} & 
\textbf{Regularization} & 
\textbf{Initialization} &  &  \\ \hline
\ding{56} & \ding{52} & \ding{52} & \ding{52} & 0.49 & 0.78 \\ 
\ding{52} & \ding{56} & \ding{52} & \ding{52} & 0.76 & 0.81 \\ 
\ding{52} & \ding{52} & \ding{56} & \ding{52} & 0.52 & 0.73 \\ 
\ding{52} & \ding{52} & \ding{52} & \ding{56} & 0.60 & 0.76 \\ \hline
\ding{52} & \ding{52} & \ding{52} & \ding{52} & \textbf{0.46} & \textbf{0.69} \\ \hline \hline

\end{tabular}
}
\end{table}

\subsubsection{Impact of Model Size on CDDM}
To explore the impact of model size on trajectory prediction performance, we distill three CDDMs with different hidden dimensions and compare their accuracy and computational cost. As shown in Table~\ref{tab:hiddendim}, CDDM maintains consistent accuracy across model sizes, demonstrating the robustness and reliability of our method.

\subsubsection{Impact of Sample Numbers on Prediction Performance}
We evaluate the impact of the number of samples $N$ in the best-of-$N$ strategy on CDDM’s prediction performance, as shown in Table~\ref{tab:bestofn}. Increasing $N$ consistently improves accuracy, highlighting the model’s ability to generate diverse trajectory samples. However, larger $N$ also increases sequential inference time, revealing a trade-off between prediction accuracy and computational efficiency. In practice, trajectory generation can be parallelized to maintain nearly constant inference time for $N$ trajectories.

\begin{table}[h!]
\centering
\caption{Best-of-N evaluation of CDDM with $K=4$ and $K=2$ on ETH.}
\label{tab:bestofn}
\begin{tabular}{c|ccc|ccc}
\hline \hline
\multirow{2}{*}{\textbf{Best-of-N}} & 
\multicolumn{3}{c|}{$K=4$} & 
\multicolumn{3}{c}{$K=2$} \\ \cline{2-7}
 & \textbf{ADE} & \textbf{FDE} & \textbf{Inference} & \textbf{ADE} & \textbf{FDE} & \textbf{Inference} \\ \hline
1  & 1.43 & 2.82 & $\sim$9 ms & 1.24 & 2.37 & $\sim$6 ms\\ 
5  & 0.71 & 1.28 & $\sim$35 ms & 0.71 & 1.26 & $\sim$20 ms\\ 
10 & 0.58 & 0.94 & $\sim$69 ms & 0.59 & 0.98 & $\sim$37 ms\\ 
20 & 0.46 & 0.69 & $\sim$134 ms & 0.49 & 0.74 & $\sim$67 ms \\ 
40 & 0.42 & 0.52 & $\sim$270 ms & 0.43 & 0.56 & $\sim$136 ms\\ \hline \hline
\end{tabular}
\end{table}

\subsubsection{Impact of Sampling Steps on Prediction Accuracy}
\begin{figure}[h!]
    \centering
    \includegraphics[width=\columnwidth]{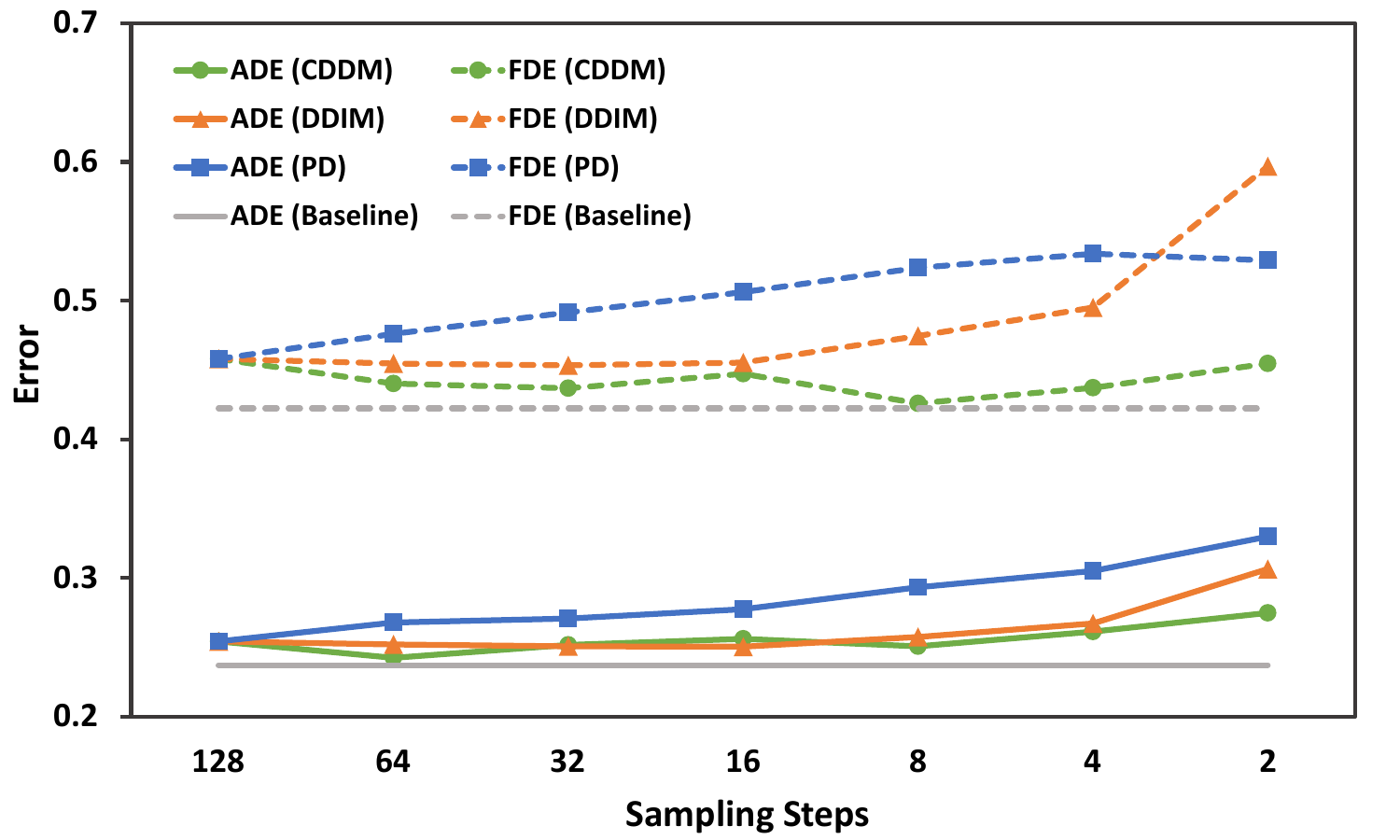}
    \caption{Impact of sampling steps on CDDM, DDIM, and PD against the baseline on ETH.}
    \label{fig:samplesteps}
\end{figure}

To investigate the impact of sampling steps on prediction accuracy, we compare our well-distilled CDDM with DDIM and PD at the lightweight model size ($H=16$) across different sampling steps $K$, as shown in Figure~\ref{fig:samplesteps}. We also include the baseline performance obtained using the original model size ($H=256$) with 128 sampling steps for reference. The results show that CDDM maintains relatively stable accuracy across distillation steps, consistently outperforming other approaches while remaining close to the baseline performance.

\subsection{Qualitative Evaluation}

In this section, we qualitatively evaluate the proposed CDDM by visualizing its trajectory prediction results for both pedestrians and vehicles. Figure~\ref{fig:vis} shows the predicted scenes produced by a well-distilled lightweight CDDM using only 4 sampling steps. We visualize individual predicted trajectories to highlight sampling diversity, and plot their joint distributions to present prediction confidence. As demonstrated in the plots, for pedestrians, CDDM effectively captures highly heterogeneous trajectory patterns and models dynamic pedestrian behavior. It is capable of generating diverse predictions, thereby accounting for the stochastic nature of pedestrian intentions. In some cases, the model even successfully predicts sudden changes in walking direction. For vehicles, CDDM accurately predicts turns and travel distances while considering interactions with surrounding agents, thereby producing precise and smooth trajectories.


\section{Conclusion}
\label{sec:conclusion}
In this paper, we propose Collaborative-Distilled Diffusion Models (CDDM), a novel method for accelerated and lightweight trajectory prediction by jointly addressing two key challenges in deploying diffusion models: slow inference speed and large model size. Built upon our Collaborative Progressive Distillation (CPD) framework, CDDM progressively transfers knowledge from a large, high-capacity teacher model to a lightweight student model, collaboratively reducing both the sampling steps and the model size in each distillation iteration.

Unlike existing approaches that focus on either acceleration or compression in isolation, our framework unifies both objectives and introduces a dual-signal regularized distillation loss that prevents overfitting and ensures consistent performance across distillation iterations. Through extensive experimental evaluation on the ETH-UCY pedestrian benchmark and the nuScenes vehicle benchmark, CDDM demonstrates its state-of-the-art prediction accuracy while significantly reducing model size, computational cost and inference latency. The well-distilled lightweight CDDM requires only 4 or 2 sampling steps, runs in 9 ms, and contains just 231K parameters, making it well-suited for real-time application on resource-constrained edge devices.

The qualitative results further highlight CDDM’s ability to generate diverse and accurate trajectories, even under dynamic agent behaviors and complex social interactions. By bridging the gap between high-performing generative models and real-world deployment constraints, CDDM sets a promising direction for future research in resource-efficient probabilistic prediction for autonomous driving and intelligent transportation systems.

\bibliographystyle{IEEEtran}
\bibliography{reference}

\vfill

\end{document}